\documentclass[conference]{IEEEtran}
\usepackage[english]{babel}

\usepackage{amsmath} 
\usepackage{graphicx}
\usepackage[colorlinks=true, allcolors=blue]{hyperref}
\usepackage{amssymb}
\usepackage[flushleft]{threeparttable}
\usepackage{array,booktabs,makecell}
\usepackage[font=small]{caption}
\usepackage{flushend}

\title{Deep Convolutional Autoencoder for Assessment of Drive-Cycle Anomalies in Connected Vehicle Sensor Data}

\author{
\IEEEauthorblockN{
    Anthony Geglio \IEEEauthorrefmark{1} \IEEEauthorrefmark{2}, 
    Eisa Hedayati \IEEEauthorrefmark{3} \IEEEauthorrefmark{2}, 
    Mark Tascillo\IEEEauthorrefmark{4}, 
    Dyche Anderson\IEEEauthorrefmark{4}, 
    Jonathan Barker\IEEEauthorrefmark{4}, 
    Timothy C.~Havens\IEEEauthorrefmark{1}}\\
	\IEEEauthorblockA{
	\IEEEauthorrefmark{1} Computer Science Department\\
	                    Michigan Technological University \\
	                    Houghton, MI 49930 \\
	                    \texttt{ajgeglio@mtu.edu, thavens@mtu.edu}} \\

	\IEEEauthorblockA{
    \IEEEauthorrefmark{3} Department of Radiology\\
	                    New York University School of Medicine\\
	                    New York, NY 10016 \\
	                    \texttt{eisa.hedayati@nyu.edu}\\
	                    \IEEEauthorrefmark{2} Denotes equal contribution} \\
    \IEEEauthorblockA{
	\IEEEauthorrefmark{4} Ford Motor Company\\
	                        Dearborn, MI 48126}}

\begin{document}
\maketitle

\begin{abstract}
This work investigates a practical and novel method for automated unsupervised fault detection in vehicles using a fully convolutional autoencoder. The results demonstrate the algorithm we developed can detect anomalies which correspond to powertrain faults by learning patterns in the multivariate time-series data of hybrid-electric vehicle powertrain sensors. Data was collected by engineers at Ford Motor Company from numerous sensors over several drive cycle variations. This study provides evidence of the anomaly detecting capability of our trained autoencoder and investigates the suitability of our autoencoder relative to other unsupervised methods for automatic fault detection in this data set. Preliminary results of testing the autoencoder on the powertrain sensor data indicate the data reconstruction approach availed by the autoencoder is a robust technique for identifying the abnormal sequences in the multivariate series. These results support that irregularities in hybrid-electric vehicles' powertrains are conveyed via sensor signals in the embedded electronic communication system, and therefore can be identified mechanistically with a trained algorithm. Additional unsupervised methods are tested and show the autoencoder performs better at fault detection than outlier detectors and other novel deep learning techniques. 

\end{abstract}

\section{Introduction}

The modern passenger vehicle has undergone major advancements in recent years due the demand for high-tech functionality, which is accompanied by a growing number of interconnected electronic components and a respective stream of sensor data. This data availability coupled with industry competition creates the need for intelligent information systems and design. Vehicle sensor data has applications for the advancement of autonomous driving security \cite{zhou_anomaly_2019,novikova_autoencoder_2020}, driver behavior analysis \cite{abdennour_driver_2021,Yadav_2019}, fuel consumption and efficiency \cite{yeh_using_2019,Qi_2017}, road object detection \cite{Wang_Hai_2019}, and sensor fusion for autonomous vehicle perception \cite{Fayyad_2020}. For vehicle manufacturers, computational intelligence is also applied to improve quality in areas such as production, logistics, and vehicle fleet management \cite{sayedahmed_computational_2021}.

The focus of this work is to explore methods of automatic anomaly detection to distinguish rare and abnormal temporal patterns in embedded vehicle sensor data which in turn can be used for fault detection. The data are from several powertrain components interconnected in the vehicle's electronic control modules and collected by engineers at Ford Motor company during drive cycle testing. The drive cycles refer to a standardized test methodology used by vehicle manufacturers to gauge various performance metrics of new vehicles by simulating a variety of typical driving conditions \cite{eva_2000}. This work was commissioned as a key component in the development of early fault detection systems in hybrid-electric vehicle fleets. The research looks beyond the traditional diagnostics of on-board systems which monitor a limited sensor domain related to emissions control. On-board diagnostics have been an important component of passenger vehicle functionality for decades and have lead to standardized monitoring and communications protocols such as ISO 9141 and 15031 \cite{ISO9141, ISO15031}. The data we analyze, which is composed of electronic signals from several powertrain components during drive-cycles, have been less utilized for diagnostics. By analyzing the stream of data from 56 sensors related to the hybrid-electric vehicles' torque and power distribution, we explore a less prescriptive but potentially more holistic vehicle health assessment that could extend sensor diagnostics in smart vehicles.

For the task of automatic fault detection, we propose an anomaly detector which utilizes a convolutional deep artificial neural network encoder-decoder, i.e., a convolutional autoencoder. Our results show that by training the network on normal drive cycle samples, our autoencoder is capable of clearly distinguishing anomalous samples that correlate to documented drive cycle faults. We contrast this method to three types of unsupervised outlier detection algorithms which can be implemented quickly with shorter training time, however are not able to clearly differentiate the data sets. The automatic detection of faults using the powertrain sensor data has a multitude of practical applications which all belong to the emerging applications of computational intelligence in modern engineered systems.

\subsection{Related work}

The subject of anomaly detection in time-series data is a vast domain due breadth of applications that are characterized by time series, and the unique complex patterns inherent for a particular problem. One comprehensive review of time-series anomaly detection describes it as "the anomaly detection wilderness" where the authors explain in their study of over 70 algorithms that selecting a suitable algorithm for a particular anomaly detection task is extremely difficult \cite{Schmidl_2022}. The authors describe six major "families" of anomaly detection algorithms which are categorized as: forecasting, reconstruction, distance, encoding, distribution, and tree methods, where the use of autoencoders falls into the reconstruction family anomaly detection techniques. The authors did not conclude that a single algorithms perform best, but do emphasise the advantages of simpler methods such as the generalization and lower parameter sensitivity. While more simplistic methods such as the distance-based and stochastic forecasting approaches may have more generalizing ability, the reconstruction approaches, specifically that of a deep neural network autoencoder, has the ability to automatically learn a compressed representation of high dimensional data through a neural network bottleneck. This also has an advantage over other compression techniques such as PCA and matrix factorization because the non-linear relationships to the latent representation can be mapped using the activation function. \cite{aggarwal_outlier_2017}.

Traditional machine learning classification approaches to pattern recognition and fault detection, specifically of vehicle systems have been employed. Prytz et. al. \cite{Prytz437123} investigate the automated data analysis of connected vehicle sensors for fault detection and emphasize that interrelations of multiple connected signals are more likely to be indicative of abnormal conditions than individual signals. They perform supervised classification using linear regression, support vector machine, and random forest to predict faults using air-intake system data, but showed a low degree of accuracy. Theissler et. al. \cite{theissler_detecting_2017} also take a supervised approach to anomaly detection for vehicle fault prediction based on a data stream of eight vehicle sensor features related to engine temperature and control. They use an ensemble of nine traditional classifiers to detect anomalous or normal drive cycles with high classification performance. In another variation in intelligent vehicle fault detection \cite{sangha_2005}, the authors explore using a model-based technique simulating dynamic engine patterns for fault detection in the air-path of automotive engines. A supervised neural network is used to classify four groups of emissions failures. This type of approach, in contrast to ours, requires domain knowledge of the specific automotive systems and is less generalized. 

Establishing thresholds to classify outliers in high dimensional temporal data remains a unique challenge presented by the nature of the variables being studied. In the broad subject of automatic threshold and outlier classification, more generalized techniques have been proposed \cite{su_robust_2019,zhao_pyod_2019}. Although these address a similar objective, these methods are designed to classify observation level events, such as a single time-step outlier in the time-series. The temporal patterns in drive cycles, and the inter-correlations between the electronic signals over time, can provide much more information to characterize normal versus abnormal events. This is shown by the results testing the drive cycle data using the techniques developed in the \emph{Python outlier detection} (PyOD) libraries \cite{zhao_pyod_2019}. Results are shown in Figure \ref{fig:pyod}. These methods assign an anomaly score at the observation level, and the aggregated totals are used to evaluate outliers relative to training data. 


Solutions for further characterization, which enable multiple classifications of rare events, have been addressed using a semi-supervised approach by maximizing use of labels when available for small subsets of the data \cite{ruff2020deep}. The scope of this study is focused on characterization of normal and abnormal data; however, given the availability of drive cycles with various classification of system faults, similar semi-supervised approaches could be considered to characterize or differentiate such events in future work. In this work, we assume that only \emph{normal} observations are available for training the algorithm.

A novel autoencoder architecture used for anomaly detection in time-series data is proposed in  a 2019 paper using what is called a \emph{Multi-Scale Convolutional Recurrent Encoder-Decoder} (MSCRED) \cite{zhang_deep_2019}. This approach sub-samples multi-scale $nxn$ time-step sliding windows $w$ and transforms the samples using the inner product $w^{T}w$ to generate a representation of cross-sensor correlations. The encoder-decoder architecture is bridged using an attention-based convolutional \emph{long-short-term memory} (LSTM) network to capture changes in temporal patterns. The authors propose this as a method which incorporates time dependence, improves noise robustness, and interprets anomaly severity. Although shown to be successful in some use cases, the inner product representations seem to depend on the data composition of continuous and highly correlated features. The transformed inner product representations of our drive cycle data lack such a correlation, prohibiting the network from improving the reconstruction through training in our experiments. Testing indicates that our sliding window approach without feature transformation is more suited to the drive-cycle data used in this study, which is composed of uncorrelated features, highly variable ranges, and a mix of continuous and discrete features.

\subsection{Clustering-based approaches}
Notable unsupervised approaches to anomaly detection in multivariate time series also include clustering based methods using traditional proximity-based methods and \emph{Fuzzy C-Means} (FCM) clustering \cite{li_anomaly_2019,li_clustering-based_2021,aggarwal_outlier_2017,zhao_pyod_2019}. The approach in \cite{li_clustering-based_2021} proposes a clustering-based anomaly detector with specific attention to the amplitude and the shape of multivariate time series. The method employs sliding window sub-sampling, FCM clustering, and a reconstruction criteria to calculate an anomaly score. The authors then employ particle swarm optimization to improve detection. The clustering-based approach has the benefit of being less reliant on prior training data; however, drawbacks are high time and space complexity in testing which is less practical for compute-constrained, on-board diagnostics and other real-time testing applications. We now turn to describing the data used in our study.

\section{Data}
The drive cycles data set is a multi-variate time-series record of 57 electronic sensor signals of powertrain components connected via the electronic control modules in hybrid-electric vehicles. Data are recorded by test engineers who capture a wide variety of driving conditions in the drive cycles.  The electronic sensor signals are broadly categorized in Table \ref{tab:data}, and are related to the power and torque forces present in the powertrain of hybrid electric vehicles that demand a dynamic, electronically controlled torque interplay between the engine and battery powered electric motors. The data are composed mostly of continuous signal variables such as mechanical component torques, rotational speeds (RPM), and vehicle speed. Other sensor signals that represent the powertrain components are ordinal such as the transmission state (park, reverse, neutral, drive), the brake status, and battery state of charge. Raw data are composed of several drive-cycles grouped into two categories: a) drive-cycle data collected from vehicles with new batteries and normally operating powertrain components (data used for training and validation), and b) drive-cycles containing approximately 3-year old batteries with a common type of fault borne from a battery connectivity issue, and have no other notable faults in the powertrain (data unseen during training and used for testing). Both of the categories are comprised of diverse types and combinations of drive cycles that simulate different driving behaviors, decreasing any bias in the trained autoencoder related to different driving patterns. There are 271 healthy drive-cycles and 150 drive cycles with a documented fault. The median drive-cycle length is $14,000$ time-steps sampled with a frequency of 10 hertz, 1,400 seconds of observations. To generate unbiased sampling of the data set for training the network, the drive-cycles are randomly sub-sampled with a 128 time-step sliding window with random cropping of 64 samples from each drive cycle. The batch size used is 256. The data are normalized by dividing each respective feature by a known maximum value which is provided by domain experts.


\begin{table}
\caption{\label{tab:data} Summarized drive cycle data set feature categories and components}
\centering 
\begin{threeparttable}
\begin{tabular}{p{1in}|p{1.6in}}
    \textbf{Feature Category}  & \textbf{Powertrain Components}
    \\
\midrule\midrule
    Torque  &   
    \fontsize{8}{12}\selectfont Engine, electric motor, transmission, driveshaft, differentials, axles, torque request system\\
\midrule
    RPM & 
    \fontsize{8}{12}\selectfont Engine, electric motor, transmission, driveshaft, axles \\ 
\midrule
    Electric Power & 
    \fontsize{8}{12}\selectfont Battery, electric motor \\ 
\midrule
    Drive State &
    \fontsize{8}{12}\selectfont Transmission, brakes
\\
\midrule
\end{tabular}
\end{threeparttable}
\end{table}

\section{Method}

We use the method of outlier detection to develop an automatic indication that samples from a drive cycle may be expressing a patterns that differ from the learned normal behavior of typical drive cycles. The general goal of outlier detection is to separate regular and abnormal data points. As we know outlier detection is useful in many examples to remove unwanted data points, but sometimes, as in our case, we are particularly interested in the instances which deviate from normal. Similar to a cybersecurity application, we want to flag the abnormal data for further investigation.

The more traditional machine learning approaches to outlier detection rely on calculations of distance and or relative density. Distance-based approaches define an instance to be an outlier in case that it is sufficiently far from the majority of other instances in the dataset. Density-based approaches separate outliers if its density is sufficiently small relative to the average density of its neighboring instances. 

Techniques based on these approaches explore outliers in their original data spaces, or some feature decomposition like PCA, and have demonstrated to work well on linearly separable distributions with stable densities. 

 We found the most success using an unsupervised approach utilizing a deep neural network autoencoder. The unsupervised approach has advantages in real-time and operational diagnostics where faulty data is non-descriptive, and does not include labels. We distinguish the faulted drive cycle data from normal battery cycles with the non-linear representation learning capability of an autoencoder network that requires no feature selection or engineering of the signal data.  With the successful implementation, we reasonably assume that other abnormalities in drive-cycles could be distinguished using the same approach. 

The network weights are refined by regeneration of the input of normal drive cycles over numerous epochs of parameter optimization during training. The autoencoder is then tested on a held out data set of normal drive cycles not used for training (validation set) and finally the abnormal drive cycle data (the test set). The relative differences of calculated reconstruction error are analyzed for their correlation to the abnormal and normal drive cycles. 

The trained autoencoder is expected to output a reconstruction of the normal drive-cycle signal matrix with a low degree of error. After extensive training, testing is performed by regenerating a subset of held-out healthy drive cycle data. The data of interest, that should trigger some system anomaly to warrant predictive maintains, contain a set of drive-cycles by vehicles that have documented faults in the powertrains. The results of testing show the trained network can differentiate anomalous patterns in the abnormal drive-cycles through data regeneration, due to some distinct characteristics or patterns that differ from the healthy drive-cycles indicated by a vast difference in the ranges of reconstruction error. From such data, an error threshold can be determined to predict whether the drive-cycle is normal or abnormal.

\subsection{Model}

The autoencoder is a feed-forward, fully convolutional neural network with hidden layers constructed using convolutional and transposed convolutional layers. The network has no fully connected layers or dropout. The ReLU activation function is used to transfer each layer's output. The loss function computes a regularized variation between the input and reconstructed signal matrix, which are the decoder reconstructed samples of the drive-cycles. The Adam optimizer is used with an initial learning rate of $8x10^{-4}$ and a rate decay every 50 epochs of training. After numerous experiments, it was found that gradual training of the network using a step-by-step weight transfer scheme resulted in a better reconstruction of the input, and enabled the use of a deeper network while preventing run-away gradients. The approach begins by training our autoencoder with very few layers, then transferring the weights to a deeper network where training is limited to the added layers. Additional epochs are then run, finally training on the entire network for fine-tuning. 


The input dimensions of the autoencoder network is based the 58 sensor features from the hybrid-electric vehicle drive cycles which are padded to 64 for convention. The samples capture 128 time-steps which is estimated by domain experts to be a sufficient time window to capture representative patterns or snapshots of the drive cycles. The input samples to the autencoder are $X \in \mathbb{R}^{mxnx1}$ with $m = 128$ time-steps, $n=58$ features. Subsequent hidden layers are composed of 8 2-dimensional convolutional layers in the encoder mirrored by 8 transposed convolutional layers in the decoder for reconstruction of the latent signal $z \in \mathbb{R}^{2x2x128}$. The initial convolutional kernel is size $7\times 7$ followed by $3\times 3$ kernels for convolutional layers 2 through 8. The number of filters are 64, 128, 256, 512, 1024, 521, 256 and 128 for the respective layers. The output $X'$ is cropped to the original signal dimensions $58 \times 128 \times 1$ in the final reconstruction. 

\subsection{Loss Function}

The proposed method of evaluation is the relative degree of reconstruction error between normal and abnormal data. The cost function $J$ used for evaluation incorporates regularization by taking the sum of three terms: \emph{mean square error} (MSE), \emph{mean absolute error} (MAE), and the \emph{standard deviation} ($\sigma$) of the difference matrix between input and reconstructed signals shown in Equation \eqref{J_eqn}.
\begin{align} 
\nonumber J(X,X') = & \frac{1}{m} \sum_{i=0}^{m} (x^{i}-x'^{i})^{2} + \\
\nonumber & \frac{1}{m} \sum_{i=0}^{m} |x^{i}-x'^{i}| + \\
&\sum_{i=0}^{m} \sigma[|x^{i}-x'^{i}|].
\label{J_eqn} 
\end{align}
Now that we have described the autoencoder architecture and how it is used, we next present results that demonstrate its effectiveness in this application.

\section{Results}

\begin{figure*}
\centering
\includegraphics[width=0.9\linewidth]{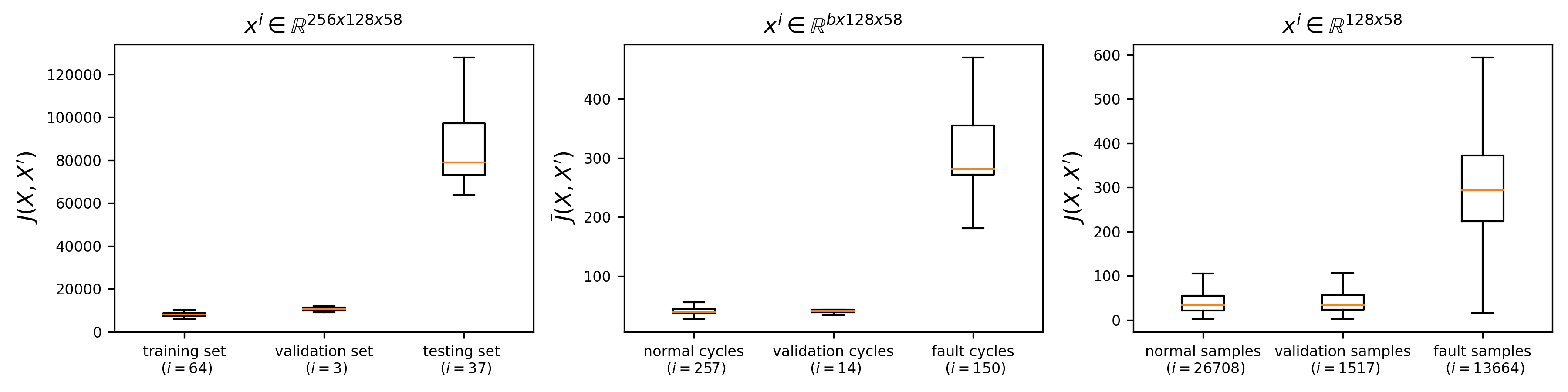}
\caption{\label{fig:exp} The reconstruction error of our trained autoencoder is shown for different sampling regimes. The left figure shows the results of training and validation reconstruction error of our autoencoder using a batch-size of 256 and 128 time-step cropped drive cycle samples. The test reconstruction error is calculated on a separate data set sampled the same way. The result shows a similar range of reconstruction error of the normal drive cycles used for training and validation, and a higher reconstruction error of the faulty drive cycles. In the middle plot, we show the average reconstruction error using the same function in Equation 1, this time testing subsequent 128 time-step samples of entire drive cycles for each of the three data sets . On the right, the plot shows the range of reconstruction error calculated per individual 128-observation sample in entire data set without the cropping used previously.  }
\end{figure*}

Figure \ref{fig:exp} show the results of testing the trained autoencoder with three different sampling regimes on the drive cycle data. The results showcase the test reconstruction error using the cost function $J(X,X')$, representing the ranges of reconstruction error: max, 75\%, median, 25\%, and min. The left plot in Fig.~\ref{fig:exp} shows the results of training and validation of the autoencoder using the cost function for the batches of cropped data used to train the network $x^i \in \mathbb{R}^{256x128x58}$. Cropping of the data set created an unbiased representation of the data, while reducing the data needed to training the network. To compare the reconstruction error of the training and validation data set, the test data composed of faulty drive cycles was sampled in the same fashion. The results show that the training and validation data produce a similar range of reconstruction error, while the test data set composed of abnormal drive cycles shows a much higher reconstruction error. The middle plot of Figure \ref{fig:exp} shows the averaged reconstruction error over each drive cycle sampled with a 128 time-step window $x^i \in \mathbb{R}^{bx128x58}$ without cropping the data. Each drive cycle is composed of roughly $14,000$ time-steps, therefore each drive cycle averages a batch of samples $b\approx{100}$. Finally, the plot on the left tests the reconstruction error for individual 128 time-step windows $x^i \in \mathbb{R}^{128x58}$ sampling the entirety of available data. With the first two tests, the reconstruction error is calculated over several samples, where an error threshold was able to differentiate the normal and abnormal cycles with 100\% accuracy in the test data. The third test shown on the left treats each data set as a series of individual samples of 128 time-steps and calculates the reconstruction errors of each sample which is approximately 0.8\% of a drive cycle. The reconstruction error of 26,000 samples from the normal drive cycles used in the training set, 1,500 samples from the validation set, and 13,000 samples from the faulty drive cycles demonstrates that these samples, when reconstructed with the trained autoencoder, can be classified as normal or abnormal samples with 97\% accuracy.

\begin{table}
\caption{\label{tab:AE performance}Proposed autoencoder test performance classifying the drive cycle samples based on a reconstruction error threshold with different sampling regimes}
\centering 
\begin{tabular}{p{1.4in}|p{1.0in}}
    Data grouping  & Performance
    \\
\midrule\midrule
    Random 128 time-step window cropping with batch-size 256  &   
    \fontsize{8}{12}\selectfont accuracy: 100\% \newline recall:100\% \newline F1 score: 100\%\\
\midrule
    Average reconstruction error of each drive-cycle in the three data sets & 
    \fontsize{8}{12}\selectfont accuracy: 100\% \newline recall:100\% \newline F1 score: 100\%\\ 
\midrule
    Reconstruction error distribution for all 128 time-step windows in data & 
    \fontsize{8}{12}\selectfont accuracy: 97.4\% \newline recall: 95.5\% \newline F1 score: 96.0\%\\
\midrule\midrule
\end{tabular}
\end{table}

\subsection{Model Comparison}

\begin{figure*}
\centering
\includegraphics[width=0.65\textwidth]{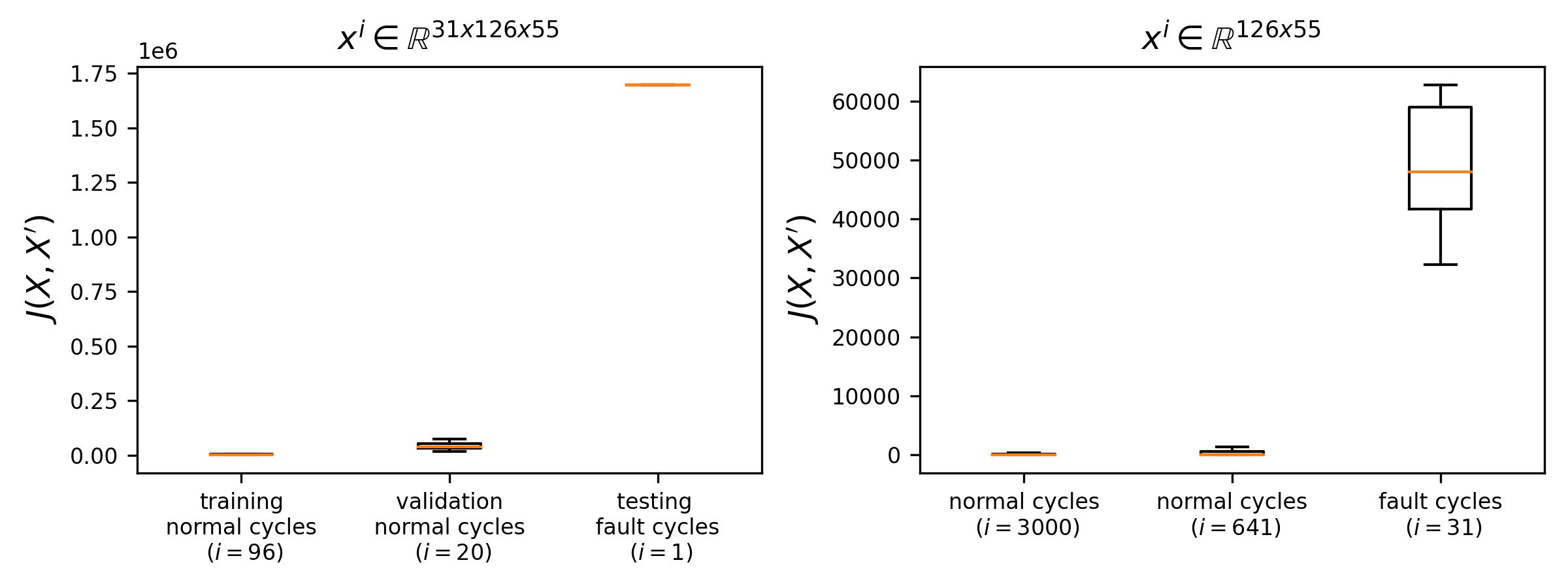}
\caption{\label{fig:kgle} Reconstruction error $J(X,X')$ calculated using our autoencoder trained on a Kaggle anomaly data set\cite{alexander_scarlat_anomaly_2021} showing results for batched data used in training (left) and re-calculated using individual samples (right).}
\end{figure*}

To test the generalizing ability of the autoencoder in its ability to distinguish abnormalities in time-series data, a comparison data set labeled with anomalous and normal temporal events is trained using the autoencoder and analyzed for the reconstruction loss between the anomalous and normal time-series events. The labeled data set is a multi-variate time series from Kaggle composed of approximately 509,000 samples with 11 features. Approximately 0.09\% of the data are classified as anomalous \cite{alexander_scarlat_anomaly_2021}. Anomalous and normal data are separated, and anomaly labeled events occur approximately once every 9-time steps. The data are broken up into chunks of 126 time-steps, evenly allocating the anomalous signals to the test batch and having dimensions appropriate for the autencoder with minimal padding. Samples were randomly shuffled and augmentation was performed to increase the feature dimension. The feature axis was replicated 5 times, randomly shuffled, and concatenated on the entire data set to expanding the 11 features to 55. Results shown in Figure \ref{fig:kgle} demonstrate the efficacy of the autoencoder to differentiate, by reconstruction error, anomalous and normal time-series data from a data set unrelated to the drive-cycles.

\begin{figure*}
\centering
\includegraphics[width=0.9\textwidth]{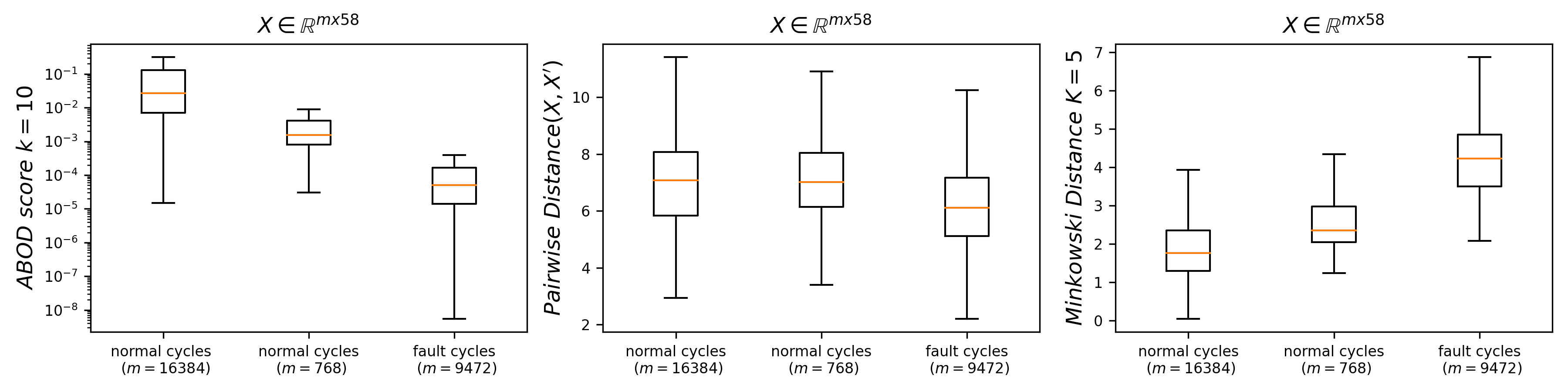}
\caption{\label{fig:pyod}  Detecting outliers in the drive cycle data set using the methods and performance metrics proposed for various classifiers in the PyOD library: ABOD (left), PyOD  autoencoder (middle), and KNN (left),  drive cycles.}
\end{figure*}

To compare the efficacy of our unsupervised deep learning approach to other approaches which require less training and are of lower network complexity, experiments were performed analyzing the drive-cycle data using a variety of approaches in the PyOD  (Python Outlier Detection) library; the angle-based anomaly detector (ABOD)\cite{kriegel_angle-based_2008} and K-nearest neighbor (KNN) outlier classifier \cite{zhao_pyod_2019}, as well as a fully connected neural network autoencoder for outlier detection in Fig.~\ref{fig:pyod}. The PyOD classifiers assign an outlier score on observation level events using the performance metric of the classifier. ABOD uses a cosine variance score relative to a specified number of nearest neighbors in euclidean space. With ABOD, a lower score indicates an outlier because a large cosine variance of surrounding neighbors indicates good clustering, wheras observations far away from the clusters have smaller cosine variances approaching zero. The KNN classifier uses a score based on the relative distance between neighbors in euclidean space, where larger distances indicate outliers. Finally, the performance metric calculated by the PyOD autoencoder calculates a pairwise distance matrix between the input and reconstructed data observations. The assumption was that if faulty drive cycles can be identified by documenting the prevalence of outliers in the data using these more basic techniques, this would negate the need for a deep neural network with high complexity and extensive training requirements. The temporal patterns and complex signal relationships, however, do not manifest at the observation level, therefore simple methods for outlier detection available in the public domain are not as well suited to characterize these patterns.

The original time-windows sampled for training, testing, and validation of the autoencoder were represented as observations by calculating an average feature vector for each sample. In other words, the sample data set that was used to train the autoencoder $X \in \mathbb{R}^{ixbxmxn}$ with i = 64 iterations of batches in the training data set, b = batch size of 256, m = 128 time-steps, and n=58 features is vertically stacked to 16,384 samples $X \in \mathbb{R}^{sxmxn}$. The results indicate that the drive-cycle data sets are not as clearly distinguishable using the outlier detectors, however the ABOD and KNN results do show a notable increase in the prevalence of outliers in the faulty drive-cycle observations relative to the normal drive cycles enabling some accuracy in classifying the observations.

The MSCRED model was tested with the drive-cycle data to compute the signal reconstruction using their proposed recurrent neural network autoencoder. MSCRED takes a similar reconstruction approach to identify anomalies in multi-variate time series data, however their study analyzes only continuous features. We performed experiments analyzing the drive cycle data set in the MSCRED model, and  after feature transformation to the so-called signature matrices, the temporal patterns in the drive-cycle sensors were not represented. The training resulted in a vanishing gradient in every experiement performed. The MSCRED algorithm was therefore unable to perform anomaly detection on our data set.

Although we did not perform a comprehensive review of all available methods, the comparison here indicates that a deep neural network reconstruction is better suited to capture the patterns in the data, and therefore classify anomalies, than distance based approaches. For our data set specifically, the following list notes the challenges faced comparing the proposed autoencoder with other models.

\begin{enumerate}
    \item There exists no direct comparison for the automotive use-case of fault detection with this variety of sensor signals; therefore, it is difficult to evaluate the relative efficacy with other fault detection and deep learning techniques applied to other sensor data because of the architecture's parameter optimization.
    \item The drive cycle features are a mix of continuous and discrete features that are binary, ordinal, and categorical features. 
    \item The dimensionality requirement of the autoencoder architecture is a limitation. Lower dimensional data can be fixed with padding, but causes over-fitting, while higher dimensional data must be reduced to the dimensionality of the architecture. Time-series data augmentation methods have been applied in other studies \cite{wen_time_2021}, however it is challenging preserve the time and space relationships between features. 
\end{enumerate}

\begin{table}
\caption{\label{tab:PYOD performance}Anomaly detector performance differentiating drive cycles based on outlier scores}
\centering 
\begin{threeparttable}
\begin{tabular}{p{1.4in}|p{1.0in}}
    Python Outlier Detection (PyOD) classifier  & Performance
    \\
\midrule\midrule
    Angle-based Outlier Detector (ABOD)\cite{zhao_pyod_2019, kriegel_angle-based_2008}  &   
    \fontsize{8}{12}\selectfont accuracy: 91.1\% \newline recall: 96.3\% \newline F1 score: 88.4\%\\
\midrule
    PyOD Autoencoder for outlier detection\cite{zhao_pyod_2019} & 
    \fontsize{8}{12}\selectfont accuracy: 60.6\% \newline recall: 36.8\% \newline F1 score: 39.6\%\\ 
\midrule
    PyOD K-nearest neighbors outlier detector\cite{zhao_pyod_2019} & 
    \fontsize{8}{12}\selectfont accuracy: 89.7\% \newline recall: 84.1\% \newline F1 score: 85.2\%\\
\midrule\midrule
\end{tabular}
\end{threeparttable}
\end{table}

\section{Conclusion}
Our results indicate that a novel approach to vehicle fault detection can be performed by analyzing multivariate time-series data from vehicle powertrain sensors connected to numerous components of hybrid-electric vehicle powertrains. The network architecture we found most suitable for the task is a fully convolutional autoencoder composed of a deep neural network of convolutional layers.  The algorithm is able to automatically capture a compressed representation of the healthy drive cycles that can be used as the basis for signal reconstruction and is successful at differentiating the drive cycles with documented faults. The autoencoder has the advantages of an unsupervised method, where it does not require a mathematical model or other prior knowledge of the normal drive-cycle patterns, and does not require identification of the specific spatial-temporal events that express the abnormalities. The autoencoder performs better at differentiating normal and abnormal vehicle drive cycles than other unsupervised techniques for outlier detection including a non-convolutional autoencoder, and an attention-based convolutional LSTM network. Further experiments also show that the model can be generalized to detect abnormalities in other time-series data non-related to drive cycles.

The automatic detection of faults using the powertrain sensor data in this study has practical applications for automotive manufactures, such as the development of new on-board diagnostics and predictive maintenance capabilities, and to improve durability testing by automotive manufacturers prior to deployment. This method for correlating abnormalities in powertrain sensor data presents a promising approach to vehicle fault detection, showcasing one of the many emerging applications where computational intelligence can be utilized to tackle the increased complexity of modern engineered systems.

\subsection{Future work}
This work demonstrated an algorithm that could distinguish anomalies in multivariate temporal drive-cycle data in hybrid-electric vehicles. In the future, we will look at how these results could be extended to consider multi-vehicle data across a connected fleet, which could further improve the ability of the autoencoder architecture to accurately discern anomalous conditions and improve methods for fleet management.

\section*{Contributions}

This paper describes work performed as part of an ongoing project advised by Dr.~Timothy Havens with funding provided by Ford Motor Company. Initial development of the code and architecture was completed by Dr.~Eisa Hedayati. Domain knowledge on the drive cycle data was provided by Ford Motor Company engineers. Supplemental experimentation including evaluation and testing of the autoencoder with additional data sets, testing of the drive-cycle data sets on other models, and review of related work was completed by Mr.~Anthony Geglio.  

\bibliographystyle{IEEEtran}
\bibliography{main}

\end{document}